\documentclass[10pt,twocolumn,letterpaper]{article}

\usepackage[pagenumbers]{cvpr} 
\usepackage[accsupp]{axessibility}

\usepackage{soul}

\usepackage[table]{xcolor}
\definecolor{mygray}{gray}{0.9}
\definecolor{mypink}{HTML}{FFEDEC}
\usepackage{multirow}

\usepackage{pifont}
\newcommand{\cmark}{\ding{51}}   
\newcommand{\xmark}{\ding{55}}   

\definecolor{cvprblue}{rgb}{0.21,0.49,0.74}
\usepackage[pagebackref,breaklinks,colorlinks,allcolors=cvprblue]{hyperref}

\def\paperID{33549} 
\def\confName{CVPR}
\def\confYear{2026}

\title{Deformation-based In-Context Learning for Point Cloud Understanding}

\author{Chengxing Lin\textsuperscript{1} \quad \space\space\space\space Jinhong Deng\textsuperscript{2} \quad \space\space\space\space Yinjie Lei\textsuperscript{3}  \quad \space\space\space\space Wen Li\textsuperscript{1,2}$^*$\\
$^1$Shenzhen Institute for Advanced Study, UESTC\\$^2$School of Computer Science and Engineering, UESTC \quad $^3$Sichuan University \\
{\tt\small linchengxing@std.uestc.edu.cn, \{jhdengvision, liwenbnu\}@gmail.com, yinjie@scu.edu.cn}
}

\begin{document}

\twocolumn[{%
    \maketitle
    \renewcommand\twocolumn[1][]{#1}
    \begin{center}
        \vspace{-0.3in}
        \captionsetup{type=figure}
        \includegraphics[width=1\textwidth]{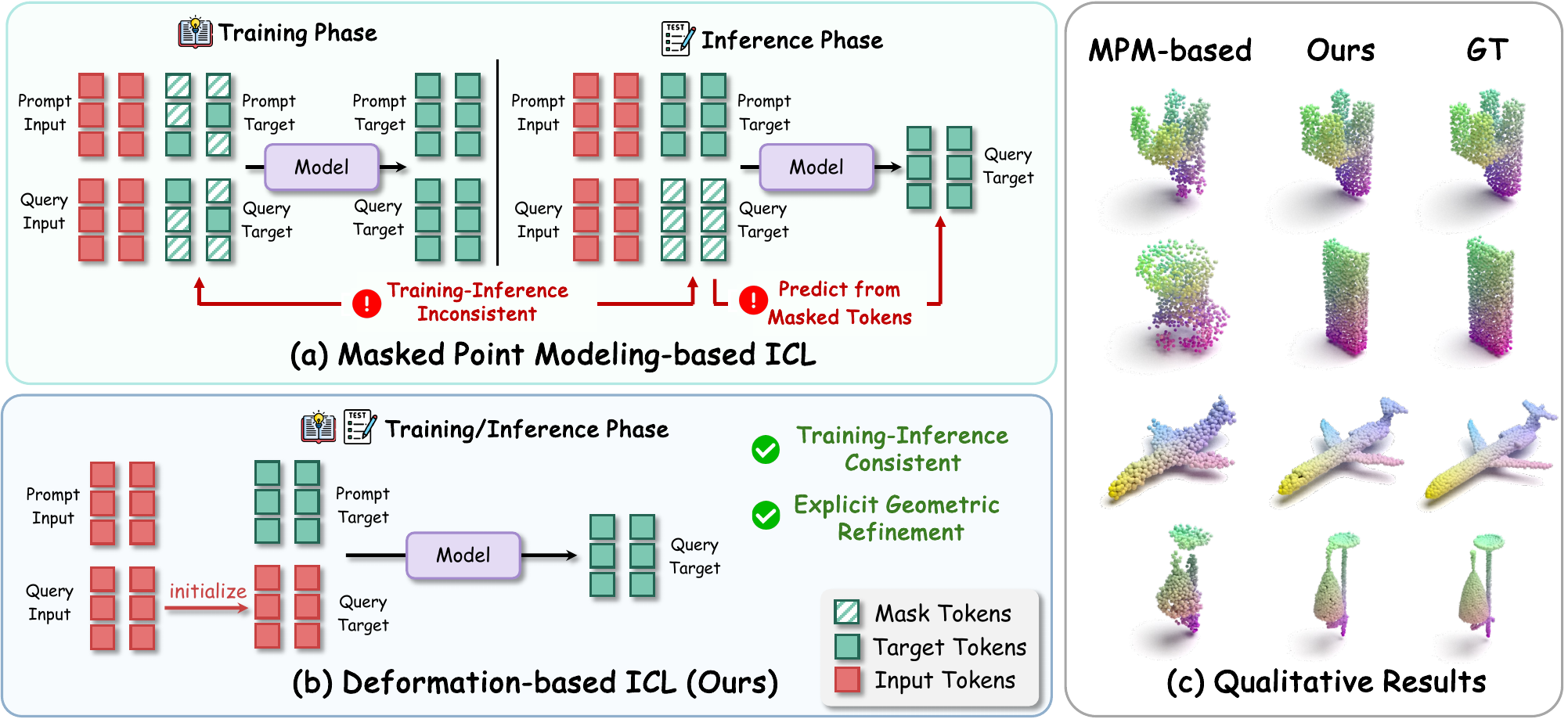}
        \caption{
            \textbf{Comparison between Masked Point Modeling (MPM)-based ICL and our Deformation-based ICL framework.}
            (a) Existing MPM methods directly predict the target from masked tokens without leveraging geometric priors, and they also suffer from a training-inference gap.
            (b) Our deformation-based ICL employs a deformation strategy to enhance geometric coherence and maintains consistency between the training and inference procedures.
            (c) Qualitative comparison on ShapeNet In-Context dataset~\cite{fang2023PIC} shows that our method predicts more accurate and complete 3D shapes than MPM-based ICL~\cite{fang2023PIC}.
        }
        \label{fig:teaser}
    \end{center}
}]

\renewcommand{\thefootnote}{\fnsymbol{footnote}}
\footnotetext{$^{*}$Corresponding author.}

\begin{abstract}
Recent advances in point cloud In-Context Learning (ICL) have demonstrated strong multitask capabilities. Existing approaches typically adopt a Masked Point Modeling (MPM)-based paradigm for point cloud ICL. However, MPM-based methods directly predict the target point cloud from masked tokens without leveraging geometric priors, requiring the model to infer spatial structure and geometric details solely from token-level correlations via transformers. Additionally, these methods suffer from a training–inference objective mismatch, as the model learns to predict the target point cloud using target-side information that is unavailable at inference time. To address these challenges, we propose \textbf{DeformPIC}, a deformation-based framework for point cloud ICL. Unlike existing approaches that rely on masked reconstruction, DeformPIC learns to deform the query point cloud under task-specific guidance from prompts, enabling explicit geometric reasoning and consistent objectives. Extensive experiments demonstrate that DeformPIC consistently outperforms previous state-of-the-art methods, achieving reductions of 1.6, 1.8, and 4.7 points in average Chamfer Distance on reconstruction, denoising, and registration tasks, respectively. Furthermore, we introduce a new out-of-domain benchmark to evaluate generalization across unseen data distributions, where DeformPIC achieves state-of-the-art performance. 
Code is available \href{https://github.com/linchengxing/DeformPIC}{here}
\end{abstract}
\vspace{-0.6cm}
\section{Introduction}
\label{sec:introduction}

3D point cloud understanding~\cite{tatarchenko2019single} plays a crucial role in a wide range of real-world applications, such as autonomous driving~\cite{hu2023uniad, li2024bevformer, chen2024end} and embodied robotics~\cite{black2024pi_0, ma2024survey_vla_embodied_ai, intelligence2025pi_05}. Despite significant advances in recent years, training a generalist model capable of handling diverse 3D point cloud tasks remains an open and challenging research problem. A promising avenue of inspiration comes from Natural Language Processing (NLP), where Large Language Models (LLMs)~\cite{touvron2023llama, touvron2023llama2, grattafiori2024llama3, guo2025deepseek, yang2025qwen3} have demonstrated impressive capabilities in In-Context Learning (ICL)~\cite{brown2020language, dong2022survey}, enabling them to generalize to unseen tasks using only a few in-context examples. Motivated by this success, recent studies~\cite{bar2022visual, wang2023images} have extended the concept of ICL from language to the visual domain, often within the Masked Image Modeling (MIM) framework~\cite{he2022mae, xie2022simmim}, leading to the development of visual in-context learning.

Building on this foundation, recent works have explored the potential of ICL for 3D point cloud understanding through Masked Point Modeling (MPM)~\cite{yu2022pointbert, pang2022point-mae, zhang2023i2p-mae}, paving the way for generalist models for 3D understanding. MPM-based methods aim to reconstruct missing patches of point clouds, thereby enabling in-context reasoning. The core idea is that training models to infer masked regions from visible contexts helps them learn to leverage prompts as contextual cues during inference. For instance, PIC~\cite{fang2023PIC} extends 2D MIM to point clouds for in-context learning and introduces the first benchmark for 3D point cloud ICL, while PIC++~\cite{liu2024PIC++} revises the in-context format for part segmentation, improving segmentation performance.


Although existing MPM-based methods~\cite{fang2023PIC, liu2024PIC++, jiang2024dg-pic, jiang2024pcotta, shao2025micas} show promising results, they still exhibit noticeable geometric artifacts. These methods face a \emph{geometry-free reconstruction} challenge, as they predict the target point clouds directly from masked tokens that carry no spatial or structural priors. These masked tokens serve solely as abstract placeholders rather than meaningful representations that encode geometric correspondences. Consequently, the model must infer spatial structure purely from token-level correlations without any explicit geometric guidance. Moreover, there exists a clear \emph{training–inference objective mismatch} in MPM-based ICL methods. During training, both the prompt and query targets are randomly masked, and the model learns to recover missing points using visible portions of both targets. In contrast, at inference time, the query target is fully masked, and the model must reconstruct the complete query geometry solely from the prompt pair and query input, without access to partial target information. This discrepancy encourages shortcut learning, where the model relies on target-side cues observed during training instead of leveraging the prompt as geometric guidance.

To explicitly model geometric correspondence and ensure consistent spatial reasoning, we propose \textbf{DeformPIC}, a deformation-based framework for point cloud ICL. Unlike previous methods that reconstruct target geometry from abstract masked tokens, DeformPIC reformulates the task as a deformation process, directly transforming the query input toward the target shape under the guidance of prompt examples. Specifically, DeformPIC decomposes point cloud ICL into two stages. First, a \textbf{Deformation Extraction Network} (DEN) extracts the geometric transformation from prompt examples, generating a task token that encodes task-specific deformation information. Second, a \textbf{Deformation Transfer Network} (DTN) applies the learned transformation to deform the query input into the target geometry. This deformation-based formulation not only enables explicit geometric manipulation but also unifies the training and inference objectives, preserving spatial continuity and ensuring geometric consistency throughout the process.

We systematically evaluate DeformPIC on both in-domain and out-of-domain benchmarks, demonstrating consistently strong performance across all settings. Specifically, we conduct experiments on the ShapeNet In-Context dataset~\cite{fang2023PIC}, which includes four point cloud understanding tasks at five difficulty levels. Additionally, we introduce a new benchmark to assess generalization across both synthetic and real-world datasets with unseen data and categories. Comprehensive experiments show that DeformPIC significantly outperforms existing state-of-the-art point cloud ICL methods across multiple datasets.

Our contributions are summarized as follows:
\begin{itemize}
\item We provide a systematic analysis of existing MPM-based point cloud ICL frameworks, highlighting the challenges of geometry-free reconstruction and the training-inference objective mismatch.
\item We introduce DeformPIC, a deformation-based framework that reformulates point cloud ICL as a deformation process, transforming the query input toward the target shape under the guidance of prompt examples.
\item We conduct extensive experiments on the ShapeNet In-Context dataset and a newly established generalization benchmark covering both synthetic and real-world datasets, showing that DeformPIC consistently outperforms state-of-the-art point cloud ICL frameworks and exhibits geometric fidelity and generalization capabilities.
\end{itemize}

\section{Related Work}
\label{sec:related_work}

\begin{figure*}[t]
    \centering
    \includegraphics[width=1.0\linewidth]{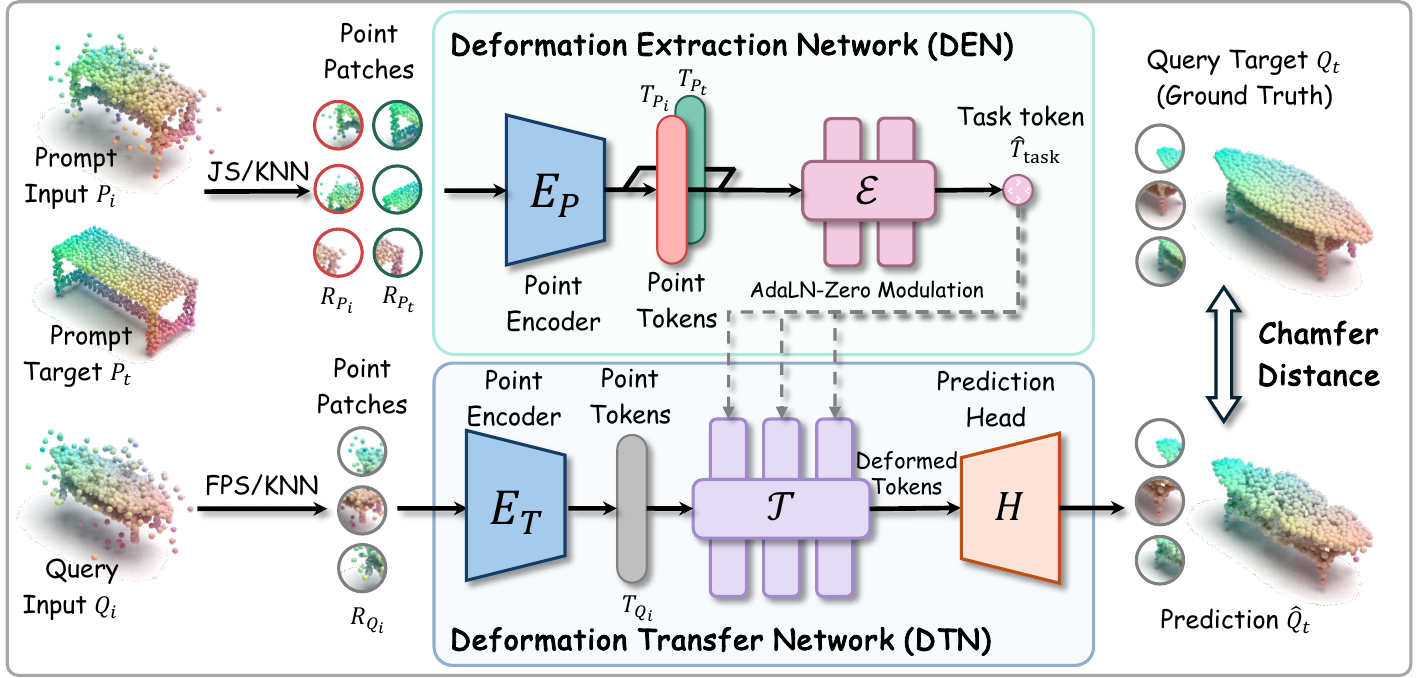}
    \caption{\textbf{Overall framework of DeformPIC.} Our model consists of two core components. The Deformation Extraction Network (DEN) extracts the geometric transformation from an example pair (prompt input\textrightarrow prompt target) into a task embedding. This embedding then modulates the Deformation Transfer Network (DTN), thereby guiding it to apply the same transformation to a new query input to produce the final prediction. The entire network is trained by minimizing the Chamfer Distance~\cite{fan2017CD} against the ground truth.}
    \label{fig:framework}
\end{figure*}

\subsection{In-Context Learning}
In-Context Learning (ICL) refers to the phenomenon where Large Language Models (LLMs) perform new tasks by conditioning on a few input–target examples supplied in the model context, without any gradient updates to model parameters. This phenomenon first emerged in GPT-3~\cite{brown2020language}. Subsequently, \cite{hendel2023context} and \cite{liu2024context} explored the underlying mechanisms of in-context learning. Both works suggest that ICL implicitly constructs task vectors from demonstration examples to guide predictions. MAE-VQGAN~\cite{bar2022visual} first brings the ICL paradigm to the field of computer vision (CV) for image inpainting tasks via training a Masked Image Modeling (MIM) model on figures of arXiv papers. Painter~\cite{wang2023images} trains MIM on the composition of input and output image pairs from various downstream tasks. Prompt Diffusion~\cite{wang2023context} conditions diffusion models on multimodal prompts of text and image pairs, unifying diverse vision-language tasks and enabling generalization to unseen tasks. 

\subsection{Point Cloud In-Context Learning}
Given the versatility of ICL in CV, researchers have extended this paradigm to 3D point cloud understanding, aiming to achieve similar generalist capabilities across diverse tasks.
PIC~\cite{fang2023PIC} first introduced the ICL paradigm to point cloud understanding via Masked Point Modeling (MPM). PIC++~\cite{liu2024PIC++} optimizes the ICL format specifically for part segmentation, demonstrating its effectiveness across multiple datasets. 
DG-PIC~\cite{jiang2024dg-pic} and PCoTTA~\cite{jiang2024pcotta} further explored point cloud ICL by leveraging transfer learning techniques to adapt models to new scenarios. SiC~\cite{wang2024SiC} and HiC~\cite{liu2025HiC} extended point cloud ICL to skeleton sequence modeling and human action modeling tasks, respectively. In contrast to these works, our method reformulates point cloud ICL by replacing masked modeling, which was originally designed for image ICL, with a deformation-based approach that is better suited for 3D point cloud data. Moreover, our approach does not rely on any transfer learning techniques or specific settings, yet achieves strong performance under cross-dataset evaluations.

\subsection{Neural Deformation}
Neural deformation aim to directly manipulate geometric structures by learning geometric deformations rather than reconstructing shapes from scratch. FlowNet3D~\cite{liu2019flownet3d} learns scene flow directly on point clouds, estimating dense deformation fields that align source and target shapes. 3D-CODED~\cite{groueix20183d_coded} introduces a template-based deformation autoencoder that maps input shapes to a canonical template mesh, enabling correspondence and reconstruction. Pixel2Mesh~\cite{wang2018pixel2mesh} progressively deforms an initial mesh into the target surface using image features, demonstrating that deformation-based strategies can preserve mesh structure during generation. Its extension, Pixel2Mesh++~\cite{wen2019pixel2mesh++}, further improves multi-view mesh reconstruction by refining deformations with cross-view consistency. These works highlight the effectiveness of directly modeling geometric transformations through deformation. In this paper, we leverage this property within the in-context learning paradigm, where deformation serves as the core mechanism to refine the query geometries under prompt guidance.
\section{Methodology}
\label{sec:method}

Instead of reconstructing target point clouds from geometry-free masked tokens, we reinterpret ICL from a geometric deformation perspective: we directly apply task-conditioned transformations to the query point cloud, where the transformations are explicitly derived from the prompt pair. As illustrated in Fig.~\ref{fig:framework}, our DeformPIC comprises two main components: the Deformation Extraction Network (DEN) and the Deformation Transfer Network (DTN). The DEN extracts task-specific information from the prompt input and target point clouds $(P_i, P_t)$ to parameterize the deformation process. The DTN then transfers the learned task-specific mapping to the query input $Q_i$, thereby producing a task-specific deformed point cloud $\hat{Q}_t$.

In the following section, we begin with a brief preliminary in Sec.~\ref{sec:preliminary}. Next, we introduce the Deformation Extraction Network (DEN) and the Deformation Transfer Network (DTN) in Sec.~\ref{sec:den} and Sec.~\ref{sec:dtn}, respectively. Finally, we present our overall training objective in Sec.~\ref{sec:objective}.

\subsection{Preliminary}
\label{sec:preliminary}

3D Point Cloud In-Context Learning (ICL) is a recently proposed paradigm for point cloud understanding that enables models to perform various tasks given a few demonstrations. Formally, each data example consists of a pair of point clouds $X = (X_i, X_t)$, where $X_i$ denotes the input and $X_t$ the corresponding target. Given a demonstration, referred to as a prompt pair $(P_i, P_t)$ that specifies the task, the model is expected to perform the same task on a query input $Q_i$ and generate the corresponding query target $Q_t$. For example, in a denoising task, both $P_i$ and $Q_i$ represent noisy inputs, while $P_t$ and $Q_t$ are their clean counterparts.

Specifically, Point-In-Context (PIC)~\cite{fang2023PIC} is the first attempt at 3D point cloud ICL, establishing both a solid baseline and a widely adopted evaluation benchmark. Specifically, given prompt pairs $(P_i, P_t)$ and query pair $(Q_i, Q_t)$, PIC uses the Joint Sampling module~\cite{fang2023PIC} to sample aligned central points between prompt data and query data, which forms prompt central points $(C_{P_i}, C_{P_t})$ and query central points $(C_{Q_i}, C_{Q_t})$, respectively. Subsequently, the K-Nearest Neighbors (KNN) algorithm queries the neighborhood points to obtain point patches $(R_{P_i},R_{P_t})$ and $(R_{Q_i},R_{Q_t})$ of these central points.

In the training phase, PIC randomly masks a portion of the target patches $(R_{P_t}, R_{Q_t})$, and replaces these patches with learnable masked tokens. Subsequently, it trains a Transformer~\cite{vaswani2017attention} model to infer the masked content from unmasked target patches and input patches $(R_{P_i}, R_{Q_i})$. PIC adopts Chamfer Distance $L_2$~\cite{fan2017CD} as the training loss, measuring the predicted masked patches $\hat R$ with its corresponding ground truth patches $R$:

\begin{equation}
    \label{eq:cdl2}
    \begin{aligned}
        \mathcal{L}(\hat R, R)=\frac{1}{|\hat R|}\sum_{p\in \hat R}\min_{g\in R}||p-g||_2^2 \\
        +\frac{1}{|R|}\sum_{g\in R}\min_{p\in \hat R}||g-p||_2^2.
    \end{aligned}
\end{equation}

In the inference phase, PIC receives the prompt pair $(P_i, P_t)$ together with the query input point cloud $Q_i$. The to-be-predicted target point cloud $Q_t$ is initialized with the learned masked tokens, which are processed by the Transformer~\cite{vaswani2017attention} together with $(P_i, P_t, Q_i)$ for reconstruction.

It is worth noting that the target point cloud is generated from masked tokens, which contain no geometric information. As a result, the model can only infer the structure and shape from the query input features through the self-attention mechanism. However, relying on such an implicit mechanism often leads to incomplete or inaccurate target geometry. Moreover, there exists a clear training–inference objective mismatch. During training, both the prompt and query targets are randomly masked, and the model learns to recover missing points using visible portions of both targets. Motivated by these limitations, we propose DeformPIC that reformulates the task as a deformation process and directly transforms the query input toward the target shape under the guidance of prompt examples. 

\subsection{Deformation Extraction Network}
\label{sec:den}
The original PIC~\cite{fang2023PIC} concatenates the prompt and query tokens and employs a single Transformer to jointly learn task semantics and geometric reconstruction. However, modeling these two inherently different objectives within a shared latent space is suboptimal. In particular, the relationship within the prompt pair (\textit{i.e.}, the task semantics) is deterministic, making cross-attention from prompt tokens to query tokens unnecessary and potentially harmful due to interference. To mitigate this, we explicitly decouple the deformation extraction from the deformation transfer process.

Specifically, the prompt pair $(P_i, P_t)$ in ICL provides rich task semantics. For example, the model must infer which task to perform (\textit{e.g.}, registration, denoising) and how to perform it (\textit{e.g.}, rotation angle, noise distribution parameters). We first employ a mini-PointNet~\cite{qi2017pointnet} as a point encoder to map point patches into point tokens:
\begin{equation}
    T_{P_i}=E_P(R_{P_i}),\quad T_{P_t}=E_P(R_{P_t}).
\end{equation}
To extract task-related features, we concatenate the prompt input tokens, prompt target tokens, and a learnable task token $T_{\text{task}}$ to form a token sequence. A stacked Transformer block, denoted by $\mathcal{E}(\cdot)$, is then applied to extract task-relevant features and aggregate them into task tokens. The output of the Transformer is a task token $\hat{T}_{\text{task}}$ that encodes the aggregated task semantics:
\begin{equation}
    \hat T_{\text{task}}=\mathcal{E}([T_{\text{task}}||T_{P_i}||T_{P_t}]),
\end{equation}
where $[\cdot||\cdot]$ represents concatenate operation.

\subsection{Deformation Transfer Network}
\label{sec:dtn}

Rather than reconstructing the target point cloud from geometry-free masked tokens, we introduce the Deformation Transfer Network (DTN), which transfers the learned deformation to the query point cloud. This deformation-based formulation realizes in-context learning by aligning the training and inference objectives.

Specifically, we encode $R_{Q_i}$ into $T_{Q_i}$ using a mini-PointNet $E_T(\cdot)$:
\begin{equation}
    T_{Q_i}=E_T(R_{Q_i}).
\end{equation}
A stack of Transformer blocks is then applied to learn deformation representations of the query input. Inspired by~\cite{peebles2023dit}, we adopt AdaLN-Zero modulation to inject the task condition into the deformation process. The overall deformation mapping can be written as:
\begin{equation}
    \hat T_{Q_t}=\mathcal{T}(T_{Q_i}, \hat T_{\text{task}}),
\end{equation}
where the $l$-th block of $\mathcal{T}(\cdot)$ can be written as:
\begin{gather}
\sigma^{(l)} = W_1^{(l)}\hat T_{\text{task}}\, \\
\eta^{(l)} = W_2^{(l)}\hat T_{\text{task}}, \\
\kappa^{(l)} = W_3^{(l)}\hat T_{\text{task}}, \\
h^{(l+1)} = h^{(l)} + \sigma^{(l)} \cdot \mathcal{A}[(1+\eta^{(l)})\cdot\text{LN}(h^{(l)}) +  \kappa^{(l)}],
\end{gather}
where $h^{(l)}$ denotes the input of the $l$-th block, LN and $\mathcal{A}$ represent layer normalization~\cite{ba2016layer_norm} and self-attention operation~\cite{vaswani2017attention}, respectively. The matrices $W_1,W_2,W_3$ are learnable MLPs whose parameters are initialized to zero. Finally, we adopt a projection head $H(\cdot)$, which is implemented as an MLP and a reshape operation, to predict point patches coordinates from point tokens:
\begin{equation}
    \hat R_{Q_t}=H(\hat T_{Q_t}).
\end{equation}

\begin{table*}[t]
\centering
\caption{\textbf{Comparison with state-of-the-art methods on the ShapeNet In-Context dataset~\cite{fang2023PIC}.} For reconstruction, denoising, and registration, we report Chamfer Distance~\cite{fan2017CD} loss ($\times 1000$). For part segmentation (Part.), we report mIoU. Since PIC~\cite{fang2023PIC} does not provide the details of specific multi-task heads, we tested with our own designed multi-task heads, which are marked with~$\dagger$. Bold: the best results. Underline: the second-best results.}
\small
\setlength\tabcolsep{0.55mm}
\renewcommand\arraystretch{1.1}
\begin{tabular}{l|c|c|cccccc|cccccc|cccccc}
\toprule[0.15em]
{\multirow{2}{*}{Models}} & {\multirow{2}{*}{Venues}} & \multicolumn{1}{c|}{Part.} & \multicolumn{6}{c|}{Reconstruction CD $\downarrow$}         & \multicolumn{6}{c|}{Denoising CD $\downarrow$}           & \multicolumn{6}{c}{Registration CD $\downarrow$}     \\
{}    & {}              & mIoU$\uparrow$     & {L1}    & L2    & L3    & L4    & L5    & {Avg.}  & L1    & L2    & L3    & L4    & L5    & {Avg.}  & L1    & L2    & L3    & L4    & L5    & {Avg.}                \\ \hline
\rowcolor{mygray}
\multicolumn{21}{c}{Task-Specific Models: trained separately for single task}           \\ \hline
\rowcolor{mygray}
{PointNet~\cite{qi2017pointnet}} & {CVPR'17}  & 77.5   & 3.7   & 3.7   & 3.8   & 3.9   & 4.1   & {3.9}   & 4.1   & 4.0   & 4.1   & 4.0   & 4.2   & {4.1}   & 5.3   & 5.9   & 6.9   & 7.7   & 8.5   & {6.9}      \\
\rowcolor{mygray}
{DGCNN~\cite{wang2019dynamic}}  & {TOG'19} & 76.1  & 3.9   & 3.9   & 4.0   & 4.1   & 4.3   & {4.0}   & 4.7   & 4.5   & 4.6   & 4.5   & 4.7   & {4.6}   & 6.2   & 6.7   & 7.3   & 7.4   & 7.7   & {7.1}      \\
\rowcolor{mygray}
{PCT~\cite{guo2021pct}}  & {CVM'21} & 79.5 & 2.4   & 2.4   & 2.5   & 2.6   & 3.0   & {2.6}   & 2.3   & 2.2   & 2.2   & 2.2   & 2.3   & {2.2}   & 5.3   & 5.7   & 6.3   & 6.9   & 7.2   & {6.3}         \\
\rowcolor{mygray}
{ACT~\cite{dong2022act}}  & {ICLR'23}  & 81.2 & 2.4   & 2.5   & 2.3   & 2.5   & 2.8   & {2.5}   & 2.2   & 2.3   & 2.2   & 2.3   & 2.5   & {2.3}   & 5.1   & 5.6   & 5.9   & 6.0   & 7.0   & {5.9}         \\ \hline
\multicolumn{21}{c}{Multitask Models: shared backbone + multiple task heads}               \\ \hline
{PointNet~\cite{qi2017pointnet}}   & {CVPR'17} & 15.3  & 87.2  & 86.6  & 87.3  & 90.8  & 92.2  & {88.8}  & 17.8  & 22.0  & 25.6  & 30.4  & 33.2  & {25.8}  & 25.4  & 22.6  & 24.9  & 25.7  & 26.9  & {25.1}        \\
{DGCNN~\cite{wang2019dynamic}}  & {TOG'19} & 17.0  & 38.8  & 36.6  & 37.5  & 37.9  & 42.9  & {37.7}  & 6.5   & 6.3   & 6.5   & 6.4   & 7.1   & {6.5}   & 12.5  & 14.9  & 17.9  & 19.7  & 20.7  & {17.1}        \\
{PCT~\cite{guo2021pct}}  & {CVM'21} & 16.7  & 34.7  & 44.1  & 49.9  & 50.0  & 52.3  & {46.2}  & 11.2  & 10.3  & 10.7  & 10.2  & 10.5  & {10.6}  & 24.4  & 26.0  & 29.6  & 32.8  & 34.7  & {29.5}        \\ \hline
\multicolumn{21}{c}{Multitask Models: pre-trained shared backbone + multiple task heads}    \\ \hline
{Point-MAE~\cite{pang2022point-mae}} & {ECCV'22}& 5.4  & 5.5  & 5.5  & 6.1  & 6.4  & 6.4  & {6.0}  & 5.6  & 5.4  & 5.6  & \underline{5.5}  & \underline{5.8}  & {5.6}  & 11.4  & 12.8  & 14.8  & 16.0  & 16.9  & {14.5}        \\ 
{ACT~\cite{dong2022act}} & {ICLR'23} & 12.1 & 7.4  & 6.6  & 6.5  & 6.6  & 7.0  & {6.8}  & 7.3  & 6.8  & 7.0  & 6.8  & 7.2  & {7.0}  & 12.2  & 14.4  & 19.4  & 25.5  & 29.0  & {20.1}        \\ 
{I2P-MAE~\cite{zhang2023i2p-mae}} & {CVPR'23} & 22.6 & 17.0  & 16.0  & 16.7  & 17.2  & 18.5  & {17.2}  & 20.6  & 20.4  & 20.1  & 18.3  & 18.8  & {19.6}  & 32.5  & 31.3  & 31.1  & 31.6  & 31.2  & {31.5}        \\ 
{ReCon~\cite{qi2023recon}} & {ICML'23} & 7.7 & 12.4  & 12.1  & 12.4  & 12.5  & 13.1  & {12.5}  & 20.4  & 24.5  & 27.2  & 29.2  & 32.5  & {26.9}  & 14.7  & 16.3  & 19.2  & 21.5  & 22.5  & {18.8}        \\
{PCP-MAE$^\dagger$~\cite{zhang2024pcp-mae}} & {NIPS'24}   & 49.8  & 4.0  & 4.1  & 4.4  & 4.6  & 6.0  & {4.6}  & 7.6  & 8.4  & 9.1  & 9.8  & 10.7  & {9.1}  & 17.3  & 25.6  & 36.3  & 45.1  & 50.2  & {34.9}        \\
{UniPre3D$^\dagger$~\cite{wang2025unipre3d}} & {CVPR'25}  & 44.0  & 3.9  & 4.0  & 4.2  & \underline{4.3}  & 5.6  & {4.4}  & 10.3  & 10.9  & 11.6  & 12.2  & 13.3  & {11.6}  & 18.2  & 26.3  & 36.5  & 44.9  & 49.7  & {35.1}    \\ \hline
\multicolumn{21}{c}{In-Context Learning Models: task-agnostic architecture for multiple tasks}                     \\ \hline
{PIC-Cat~\cite{fang2023PIC}}   & {NIPS'23}   & 79.0   & \underline{3.2}   & \underline{3.6}   & 4.6   & 4.9   & \underline{5.5}   & \underline{4.3}   & \underline{3.9}   & \underline{4.6}   & \underline{5.3}   & 6.0   & 6.8   & \underline{5.3}   & 10.0  & 11.4  & 13.8  & 16.9  & 18.6  & {14.1}         \\
{PIC-Sep~\cite{fang2023PIC}}  & {NIPS'23}    & 75.0   & 4.7   & 4.3   & 4.3   & 4.4   & 5.7   & {4.7}   & 6.3   & 7.2   & 7.9   & 8.2   & 8.6   & {7.6}   & 8.6   & 9.2   & 10.2  & 11.3  & 12.4  & {10.3}         \\ 

{PIC-S-Cat~\cite{liu2024PIC++}}   & {ArXiv'24}   & \underline{83.8}   & 9.3  & 5.1  & 4.8  & 5.0  & 10.3  & {6.9}  & 4.7  & 5.7  & 6.5  & 7.4  & 8.2  & {6.5}  & 12.8  & 15.8  & 23.9  & 31.2  & 36.9  & {24.1}    \\
{PIC-S-Sep~\cite{liu2024PIC++}}  & {ArXiv'24}    & 83.7  & 4.6  & 4.5  & 4.5  & 4.8  & 7.1  & {5.1}  & 9.4  & 11.7  & 12.5  & 13.1  & 13.4  & {12.0}  & \underline{6.0}  & \underline{6.1}  & \underline{7.6}  & \underline{6.7}  & \underline{7.3}  & \underline{6.7}    \\
\rowcolor{mypink}
{\textbf{DeformPIC}}  & {Ours}    & \textbf{83.9}   & \textbf{2.2}  & \textbf{2.3}  & \textbf{2.5}  & \textbf{2.8}  & \textbf{3.9}  & {\textbf{2.7}}  & \textbf{2.8}  & \textbf{3.3}  & \textbf{3.6}  & \textbf{3.8}  & \textbf{3.9}  & {\textbf{3.5}}  & \textbf{1.9}  & \textbf{1.9}  & \textbf{1.9}  & \textbf{2.0}  & \textbf{2.1}  & {\textbf{2.0}}       \\ \bottomrule
\end{tabular}
\label{tab:main_result}
\vspace{-1em}
\end{table*}

\subsection{Overall Training Objective}
\label{sec:objective}

A common neural deformation objective~\cite{jiang2020shapeflow, tang2022neural_deformation} is to predict a displacement field over points. However, in ICL settings, large variations in local displacements can hinder optimization stability and convergence. Therefore, we adopt an end-to-end deformation objective that directly predicts the deformed point cloud. Recalling Eq.~(\ref{eq:cdl2}), we use the $L_2$ variant of the Chamfer Distance as the loss function to optimize the entire network, formulated as:
\begin{equation}
    \begin{aligned}
    \mathcal{L}(\hat R_{Q_t},R_{Q_t})=\frac{1}{|\hat R_{Q_t}|}\sum_{p\in \hat R_{Q_t}}\min_{g\in R_{Q_t}}||p-g||_2^2 \\
    +\frac{1}{|R_{Q_t}|}\sum_{g\in R_{Q_t}}\min_{p\in \hat R_{Q_t}}||g-p||_2^2.
\end{aligned}
\end{equation}


Unlike previous MPM-based ICL frameworks, such as those in~\cite{fang2023PIC, liu2024PIC++, jiang2024dg-pic}, which introduce a mismatch between training and inference by reconstructing masked regions during training and performing full prediction during inference, DeformPIC maintains a consistent deformation objective throughout both phases. By applying the same deformation process during both training and inference, DeformPIC ensures alignment between the supervision signal and the inference goal, effectively eliminating the training–inference mismatch. This design not only guarantees that the model’s geometric understanding is consistent but also preserves spatial continuity, allowing for more accurate and reliable reconstruction of point cloud geometries.
\section{Experiments}

\subsection{Experimental Setup}
\noindent\textbf{Implementation Details}
Following PIC~\cite{fang2023PIC}, we sample 1,024 points from each point cloud object. Each point cloud is then divided into 64 patches, with each patch containing 32 points. We use the AdamW optimizer~\cite{loshchilov2017fixing} with a cosine learning rate decay schedule. The learning rate is linearly warmed up from 1e-6 to 1e-4 during the first 10 epochs and then decays according to the cosine schedule. We set the weight decay to 0.05. The model is trained for a total of 300 epochs with a batch size of 128. We provide model architecture details in our supplementary. All experiments are conducted on an NVIDIA TITAN RTX 24GB GPU.

\noindent\textbf{ShapeNet In-Context Dataset.}
We evaluate DeformPIC on the ShapeNet In-Context dataset~\cite{fang2023PIC}, which is derived from ShapeNet~\cite{chang2015shapenet} and ShapeNetPart~\cite{yi2016shapenetpart}. Each sample consists of a pair of point clouds: an input to be processed and a target representing the desired output. The dataset contains 174,404 training samples and 43,050 test samples, covering four point cloud understanding tasks: reconstruction, denoising, registration, and part segmentation across the five levels of difficulty (L1-L5).

\begin{figure*}[t]
    \centering
    \includegraphics[width=1.0\linewidth]{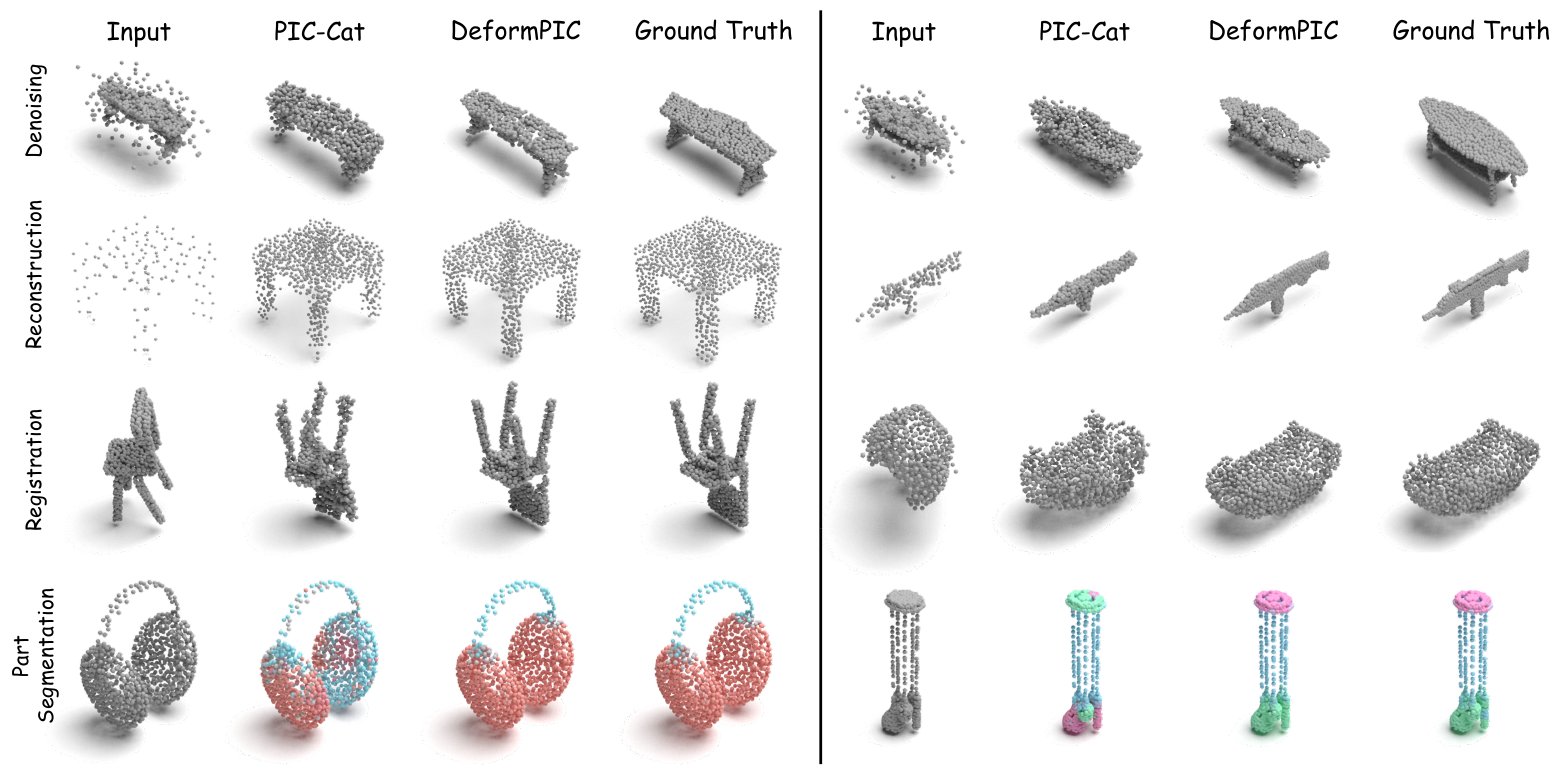}
    \vspace{-6mm}
    \caption{\textbf{Qualitative comparison on ShapeNet In-Context dataset.} From left to right: input point clouds, PIC-Cat~\cite{fang2023PIC} results, our DeformPIC results, and ground truth. Our method achieves better geometric detail preservation, improved structural coherence, and fewer visual artifacts across diverse tasks.}
    \label{fig:visualization}
\end{figure*}

\noindent\textbf{ModelNet40 / ScanObjectNN In-Context Dataset.}
To further assess the generalization capability of DeformPIC beyond ShapeNet In-Context, we construct additional in-context datasets based on ModelNet40~\cite{wu2015modelnet} and ScanObjectNN~\cite{uy2019scanobjectnn}. ModelNet40~\cite{wu2015modelnet} dataset consists of 12,311 clean, synthetic CAD models covering 40 object categories. ScanObjectNN~\cite{uy2019scanobjectnn} contains about 15,000 real-world scanned objects with background noise and occlusions. This allows us to comprehensively evaluate DeformPIC under both ideal and challenging real-world conditions.
For each dataset, we followed the same procedure as for ShapeNet In-Context to construct paired samples, where each sample consists of an input point cloud and a corresponding target output for a specific task.
More details about the dataset construction can be found in our supplementary.
We respectively denote the processed ModelNet40 and ScanObjectNN datasets as ModelNet40 In-Context and ScanObjectNN In-Context.

\noindent\textbf{Baseline Methods.}
Task-specific models employ standard point cloud backbones (PointNet~\cite{qi2017pointnet}, DGCNN~\cite{wang2019dynamic}, PCT~\cite{guo2021pct}, ACT~\cite{dong2022act}) trained independently for each task, representing fully specialized models without cross-task sharing.
Multitask models share a backbone with multiple task heads and are trained jointly on multiple tasks. For comparison, we include both conventional point cloud backbones~\cite{qi2017pointnet, wang2019dynamic, guo2021pct} and recent self-supervised pretraining methods~\cite{pang2022point-mae, dong2022act, zhang2023i2p-mae, qi2023recon, zhang2024pcp-mae, wang2025unipre3d}.
ICL models are trained following the point cloud ICL paradigm. We compare against the canonical MPM-based PIC~\cite{fang2023PIC} and the enhanced PIC-S~\cite{liu2024PIC++}, which reformulates part segmentation format within the point cloud ICL framework for improved dense prediction performance.

\subsection{In-domain Benchmark}
We evaluate DeformPIC under the in-domain setting, where both the training and testing samples come from the same data distribution. All experiments are conducted on the ShapeNet In-Context dataset~\cite{fang2023PIC}.

\noindent\textbf{Main Results.} We present the main results in Table~\ref{tab:main_result}. We observe that DeformPIC outperforms both the PIC series~\cite{fang2023PIC} and the PIC-S series~\cite{liu2024PIC++} across all tasks compared with previous ICL models.
Specifically, on the registration task, our method achieves an average performance of 2.0 CD, significantly surpassing the previous state-of-the-art PIC-S-Sep (6.7 CD). This improvement is also evident in the reconstruction and denoising tasks, where DeformPIC outperforms PIC-Cat by 1.6 and 1.8 CD, respectively. Notably, DeformPIC also achieves better part segmentation accuracy than the PIC-S series~\cite{liu2024PIC++}, despite the latter being specifically re-designed for this task.

Compared to multitask models, DeformPIC consistently outperforms state-of-the-art methods across all four tasks. For reconstruction, DeformPIC surpasses UniPre3D~\cite{wang2025unipre3d} at all five difficulty levels, with an average improvement of 1.7. For denoising, it exceeds the best-performing Point-MAE~\cite{pang2022point-mae} by an average of 2.1. For registration, DeformPIC achieves a substantial gain (2.0 \textit{v.s.}14.5) over Point-MAE~\cite{pang2022point-mae}. Furthermore, it delivers a remarkable 34.1 mIoU improvement over PCP-MAE~\cite{zhang2024pcp-mae} on this task. 
When compared to task-specific models, DeformPIC excels in part segmentation, easier reconstruction tasks (L1 and L2 levels), and registration. Notably, for registration, DeformPIC outperforms the specialized ACT~\cite{dong2022act} (2.0 \textit{v.s.} 5.9).

Overall, DeformPIC outperforms all previous point cloud ICL approaches, validating the crucial role of aligning training and inference objectives. Its superior registration performance is a direct result of leveraging the geometric priors of the query point cloud through deformation, rather than relying on abstract masked tokens for shape reconstruction. This approach preserves the structural integrity of the point cloud, ensuring more accurate and consistent geometric reasoning throughout both training and inference.

\begin{table}[t]
\centering
\caption{\textbf{Comparison of generalization ability with state-of-the-art methods on the ModelNet40 In-Context (M.) and ScanObjectNN In-Context (S.) datasets.} We report task-averaged Chamfer Distance~\cite{fan2017CD} ($\times 1000$).}
\setlength\tabcolsep{0.95mm}
\begin{tabular}{l|cc|cc|cc}
\toprule[0.15em]
{\multirow{2}{*}{Models}}  & \multicolumn{2}{c|}{Rec. CD $\downarrow$}  & \multicolumn{2}{c|}{Den. CD $\downarrow$}  & \multicolumn{2}{c}{Reg. CD $\downarrow$}     \\
{}                   & {M.}  & {S.} & {M.}  & {S.} & {M.}  & {S.} \\ \hline
\multicolumn{7}{c}{Multitask Models}    \\ \hline
{Point-MAE~\cite{pang2022point-mae}}   & \underline{5.0} & \underline{5.3}  & \underline{8.1} & {9.6}  & {38.2} & {26.1}     \\
{ReCon~\cite{qi2023recon}}              & {5.7} & {5.8}  & {9.5} & {10.9}  & {37.9} & {41.9}       \\
{PCP-MAE~\cite{zhang2024pcp-mae}}      & {5.7} & {6.0}  & {9.8} & {11.2}  & {37.6} & {42.8}    \\
{UniPre3D~\cite{wang2025unipre3d}}     & {5.5} & {5.6}  & {12.6} & {14.8}  & {38.7} & {42.9}      \\ \hline
\multicolumn{7}{c}{In-Context Learning Models}                     \\ \hline
{PIC-Cat~\cite{fang2023PIC}}         & {7.4} & {10.6}  & \underline{6.9} & \underline{9.2}  & {16.2} & {20.2}      \\
{PIC-Sep~\cite{fang2023PIC}}         & {6.3} & {6.1}  & {12.6} & {11.4}  & \underline{8.9} & \underline{9.0}   \\
\rowcolor{mypink}
{\textbf{DeformPIC}}    & \textbf{3.4} & \textbf{4.0}  & \textbf{3.9} & \textbf{5.0}  & \textbf{2.3} & \textbf{2.0}   \\
\bottomrule[0.1em]
\end{tabular}
\label{tab:transfer_result}
\vspace{-1em}
\end{table}

\noindent\textbf{Qualitative Results.}
We present qualitative comparisons in Fig.~\ref{fig:visualization}.
DeformPIC produces results that are more consistent with both the overall shape and fine-grained geometric details of the input point clouds across multiple tasks.
For reconstruction and denoising, our model generates more reasonable outputs than PIC-Cat~\cite{fang2023PIC}, although some subtle discrepancies remain compared to the ground truth.
For registration, since the input point clouds already provide complete geometry, our direct deformation strategy effectively preserves both global structure and local details, yielding predictions that closely match the ground truth.
For part segmentation, while PIC-Cat~\cite{fang2023PIC} struggles to delineate certain semantic parts, DeformPIC performs well on these challenging regions, demonstrating robust performance. We present more qualitative results in our Supplementary Materials.

\begin{table}[t]
    \centering
    \renewcommand\arraystretch{1}
    \setlength\tabcolsep{5pt}
    \caption{\textbf{Ablation Study of DeformPIC on ShapeNet In-Context.} We report the task-averaged Chamfer Distance ($\times 1000$) for the reconstruction, denoising, and registration tasks.}
    \scalebox{1.0}{
    \begin{tabular}{c|ccc|ccc}
        \toprule[0.15em]
        Model & Consis. & DTN & DEN & Rec. & Den. & Reg. \\
        \midrule
        \multirow{2}{*}{PIC-Cat}  & \xmark & \xmark & \xmark & 4.3 & 5.3 & 14.1 \\
          & \cmark & \xmark & \xmark & 3.2 & 4.0 & 3.5 \\
        \cline{1-1}
        \multirow{2}{*}{Ours}       & \cmark & \cmark & \xmark & 3.1 & 3.8 & 2.9 \\
                                   & \cellcolor{mypink}\cmark & \cellcolor{mypink}\cmark & \cellcolor{mypink}\cmark
                                   & \cellcolor{mypink}\textbf{2.7} & \cellcolor{mypink}\textbf{3.5} & \cellcolor{mypink}\textbf{2.0} \\
        \bottomrule[0.1em]
    \end{tabular}}
    \label{tab:ablation_main}
\vspace{-1em}
\end{table}

\begin{figure*}[t]
    \centering
    \includegraphics[width=\textwidth]{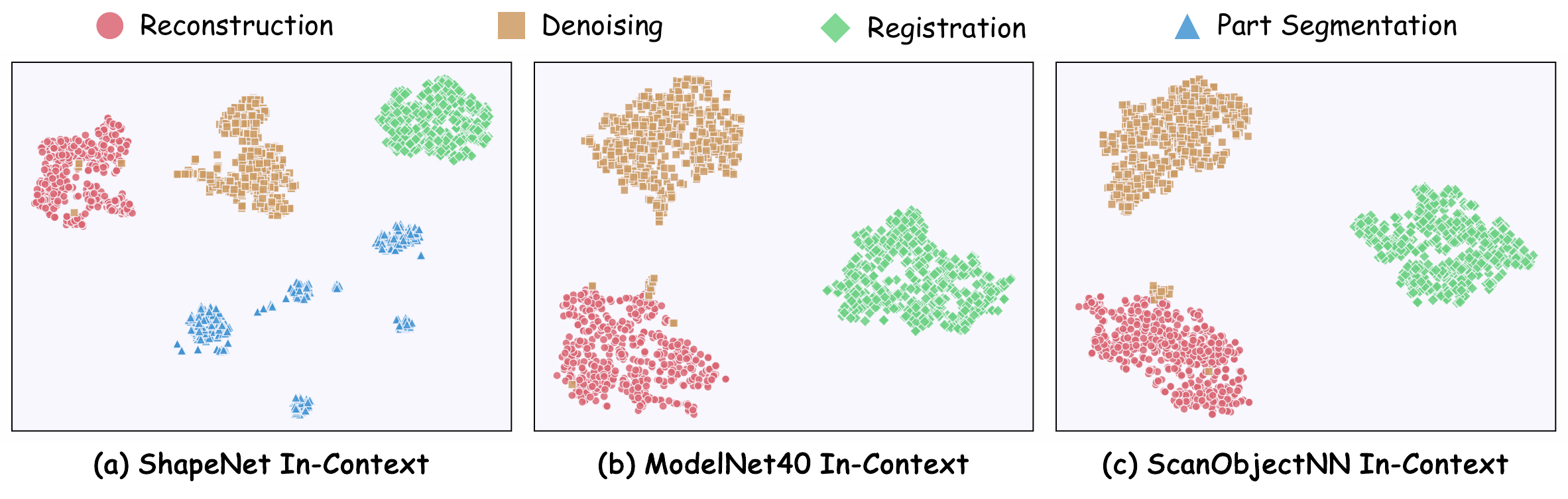}
    \caption{\textbf{Task Feature Visualization of DeformPIC on the ShapeNet / ModelNet40 / ScanObjectNN In-Context datasets.} We use t-SNE~\cite{maaten2008tsne} to reduce the dimensionality of the task features to a 2D space and visualize the distributions of task features.}
    \label{fig:tsne}
\end{figure*}

\subsection{Out-of-domain Benchmark}
To evaluate the generalization ability of DeformPIC under domain shift, we propose an out-of-domain (OOD) benchmark. In this benchmark, the model is trained on the ShapeNet In-Context dataset~\cite{fang2023PIC} and then directly tested on ModelNet In-Context and ScanObjectNN In-Context without any fine-tuning or additional adaptation strategy. The OOD benchmark thus measures the model’s ability to adapt to novel geometric or semantic distributions purely through in-context reasoning.

\noindent\textbf{Main Results.}
The results are presented in Table~\ref{tab:transfer_result}. We observe that multitask models struggle to transfer their learned capabilities to out-of-domain data, exhibiting significant performance degradation on the denoising and registration tasks. ICL models demonstrate stronger out-of-domain generalization, which we attribute to their ability to infer from the query by following the information provided in the prompt, rather than merely learning task-specific execution patterns on the training data as multitask models do. Our DeformPIC consistently outperforms all multitask models as well as the PIC~\cite{fang2023PIC} across three tasks on two datasets.

\subsection{Experimental Analysis}
\noindent\textbf{Ablation Study.}
We conduct an ablation study on ShapeNet In-Context to analyze the contribution of each component in DeformPIC (Table~\ref{tab:ablation_main}). Starting from the baseline PIC-Cat~\cite{fang2023PIC}, we fully mask the query target while leaving the prompt pair unmasked, matching the inference setting in which the query target is masked. This yields a substantial improvement in registration performance, highlighting the importance of aligning the training-inference objective.
We then incorporate the DTN to adapt our formulation from MPM to deformation, demonstrating the value of providing explicit geometric cues. Building on this variant, we further introduce the DEN, which decouples the deformation extraction process. This component brings additional gains across all tasks, indicating that DEN strengthens the model’s ability to interpret the in-context prompts.

\noindent\textbf{Analysis of Learned Task Feature.}
To investigate whether the DEN module truly understands the task indicated by the prompt pair, we apply t-SNE~\cite{maaten2008tsne} to project the task tokens into two dimensions and visualize them, as shown in Fig.~\ref{fig:tsne}. It is evident that the task features form four well-separated clusters corresponding to the different task categories, demonstrating that the task tokens extracted by DEN effectively capture task information. Although the DEN module successfully distinguishes between tasks in most cases, we observe that a small number of samples from the denoising task are mistakenly grouped with those from the reconstruction task. This may indicate that the task representations for these two categories share overlapping features, making it challenging for the model to separate them completely. This finding suggests that further refinement of the task token representations may be needed to better distinguish between highly similar tasks.

\section{Conclusion}
In this work, we introduce DeformPIC, a novel framework for in-context learning (ICL) on point clouds. The core idea of our approach is to reformulate point cloud ICL by replacing Masked Point Modeling (MPM) with a deformation-based paradigm. Instead of predicting the target point cloud from geometry-free masked tokens, DeformPIC learns to deform the query geometry toward the target shape under the guidance of a prompt pair. DeformPIC not only enables explicit geometric manipulation but also unifies the training and inference objectives, preserving spatial continuity and ensuring geometric consistency throughout the process. We conduct extensive evaluations on both in-domain and out-of-domain benchmarks. On the in-domain benchmark, DeformPIC sets a new state of the art on the ShapeNet In-Context dataset. On the out-of-domain benchmark, DeformPIC consistently achieves state-of-the-art performance on our newly established ModelNet40 In-Context and ScanObjectNN In-Context datasets.

\section*{Acknowledgements}
This work is supported by the New Generation Artificial Intelligence-National Science and Technology Major Project (No. 2025ZD0123002), the National Natural Science Foundation of China (U23B2013 and 62276176), the Science, Technology and Innovation Project of Shenzhen Longhua District (No. 20260309G23410662) and the Shenzhen Fundamental Research Program (No.JCYJ20220530164812027).
{
    \small
    \bibliographystyle{ieeenat_fullname}
    \bibliography{main}
}

\clearpage
\setcounter{page}{1}
\maketitlesupplementary
\appendix

\section*{Overview of the Supplementary Material}
To complement the main paper, this supplementary material offers expanded implementation details to facilitate reproducibility, together with additional experiments and analyses, structured as follows:
\begin{itemize}
    \item Sec.~\ref{sec:supp_imple_detail} provides detailed descriptions of the model architectures, parameter settings, and fairness considerations in comparisons.
    \item Sec.~\ref{sec:supp_dataset_deatils} explains how the ModelNet40 and ScanObjectNN In-Context datasets are built.
    \item Sec.~\ref{sec:supp_main_result} reports additional evaluations using F-score and EMD to further demonstrate the advantages of DeformPIC.
    \item Sec.~\ref{sec:supp_more_analysis} presents additional analysis about task encoding mechanism.
    \item Sec.~\ref{sec:supp_more_visualization} presents qualitative comparisons between DeformPIC and PIC baselines across multiple tasks.
    \item Sec.~\ref{sec:supp_discussion} discusses the method’s current limitations and outlines future directions.
\end{itemize}

\section{More Implementation Details}
\label{sec:supp_imple_detail}
\noindent\textbf{Model Architecture.}
We adopt 4 transformer blocks in the DEN and 8 transformer blocks with AdaLN-Zero modulation~\cite{peebles2023dit} in the DTN by default. Both networks are configured with a hidden dimension of 384 and 6 attention heads. We apply a drop path rate of 0.1 for DTN to enhance regularization, while DEN does not employ drop path. 

\noindent\textbf{Model Comparison.}
We report detailed model comparisons in Table~\ref{tab:details_params}.  To ensure fair comparisons with baseline models, we control the model capacity by matching the total parameter count rather than the layer depth. 
Specifically, our default configuration, DeformPIC-\textit{8d}, has a comparable parameter scale to PIC-Cat and PIC-Sep (29.84M vs. 28.91M) and therefore reflects a fair capacity comparison. 
For completeness, we also implement a deeper variant, DeformPIC-\textit{12d}, whose number of layers aligns with PIC-Cat and PIC-Sep but with higher parameters introduced by AdaLN-Zero modulation. 
Both variants are trained under identical settings, and their results exhibit consistent performance trends.

\section{More Details on Dataset Construction}
\label{sec:supp_dataset_deatils}
In this section, we present the construction details of ModelNet40 In-Context and ScanObjectNN In-Context, encompassing three types of tasks (reconstruction, denoising, and registration). 
For each sample in the original dataset, we first standardize the point cloud by translating it to have zero-mean XYZ coordinates and scaling it to a unit sphere. We then generate five levels of “disruption", corresponding to increasingly challenging transformations.
\begin{itemize}
    \item \textbf{Reconstruction.} We downsample the original 1,024-point cloud to $\{512, 256, 128, 64, 32\}$ points, producing inputs of varying sparsity while keeping the original point cloud as the target..
    \item \textbf{Denoising.} We corrupt the point cloud by replacing $\{100, 200, 300, 400, 500\}$ points with Gaussian noise sampled from $\mathcal N(0, I)$.
    \item \textbf{Registration} We randomly rotate the original point cloud within angle ranges of \{±20°, ±40°, ±60°, ±80°, ±100°\}.
\end{itemize}
The full set of perturbed samples across all tasks and disruption levels constitutes our ModelNet40 In-Context and ScanObjectNN In-Context benchmarks.

\begin{table*}[t]
  \caption{Comparison of model variants in terms of computational complexity, and in-context performance in ShapeNet In-Context~\cite{fang2023PIC}.}
  \small
  \centering
  \renewcommand\arraystretch{1}
  \setlength\tabcolsep{3pt}
    \begin{tabular}{c|cc|cccccc|cccccc|cccccc}
        \toprule[0.15em]
        {\multirow{2}{*}{Models}}     & {\multirow{2}{*}{Flops(G)}}  & {\multirow{2}{*}{Param.(M)}}    & \multicolumn{6}{c|}{Reconstruction CD $\downarrow$}  & \multicolumn{6}{c|}{Denoising CD $\downarrow$}  & \multicolumn{6}{c}{Registration CD $\downarrow$}  \\ 
        {}    & {} & {}  & {L1}    & L2    & L3    & L4    & L5    & {Avg.}  & L1    & L2    & L3    & L4    & L5    & {Avg.}  & L1    & L2    & L3    & L4    & L5    & {Avg.} \\\midrule
        PIC-Cat~\cite{fang2023PIC} & 11.31   & 28.91  & 3.2   & 3.6    & 4.6  & 4.9  & 5.5 & 4.3 & 3.9 & 4.6 & 5.3 & 6.0 & 6.8 & 5.3 & 10.0 & 11.4 & 13.8 & 16.9  & 18.6 & 14.1  \\
        PIC-Sep~\cite{fang2023PIC} & 8.14    & 28.91  & 4.7   & 4.3    & 4.3  & 4.4 & 5.7  & 4.7 & 6.3 & 7.2 & 7.9 & 8.2  & 8.6 & 7.6 & 8.6 & 9.2 & 10.2 & 11.3 & 12.4 & 10.3  \\
        DeformPIC-\textit{12d}     & 5.33    & 40.48  & 2.1   & 2.2  & 2.4  & 2.7 & 3.9 & 2.7 & 2.8 & 3.3 & 3.5 & 3.7 & 3.8 & 3.4 & 1.8 & 1.8 & 1.9 & 1.9  & 2.0 & 1.9  \\
        \rowcolor{mypink}
        DeformPIC-\textit{8d}      & 4.87    & 29.84  & 2.2   & 2.3  & 2.5   & 2.8 & 3.9 & 2.7 & 2.8 & 3.3 & 3.6 & 3.8 & 3.9 & 3.5 & 1.9 & 1.9 & 1.9 & 2.0  & 2.1 & 2.0
        \\\bottomrule[0.1em]
    \end{tabular}
    \label{tab:details_params}
\end{table*}

\section{More Results}
\label{sec:supp_main_result}
To verify that our conclusions are not biased by the use of a single distance metric, we additionally report results under F-score@$\tau$~\cite{tatarchenko2019single} and Earth Mover’s Distance (EMD) on representative settings of the three in-context tasks (Reconstruction, Denoising, and Registration). The F-score@$\tau$ metric evaluates both the precision and recall of the predicted point cloud within a distance threshold $\tau$ from the ground truth.
\begin{equation}
\text{F-score@}\tau = 2 \times \frac{P(\tau) \times R(\tau)}{P(\tau) + R(\tau)},
\end{equation}
where the precision $P(\tau)$ and recall $R(\tau)$ are given by:
\begin{equation}
P(\tau) = \frac{|\{ p \in \hat{\mathcal{P}} \mid \min_{g \in \mathcal{P}} \|p - g\| < \tau \}|}{|\hat{\mathcal{P}}|},
\end{equation}
\begin{equation}
R(\tau) = \frac{|\{ g \in \mathcal{P} \mid \min_{p \in \hat{\mathcal{P}}} \|g - p\| < \tau \}|}{|\mathcal{P}|}.
\end{equation}
Here, $\hat{\mathcal{P}}$ and $\mathcal{P}$ denote the predicted and ground-truth point sets, respectively.
Intuitively, precision measures how many predicted points lie close to the real surface (i.e., accuracy of generated geometry), whereas recall measures how well the predicted point cloud covers the true surface (i.e., completeness). Thus, F-score@$\tau$ thus provides a balanced assessment of both qualities: higher values indicate more accurate and complete reconstructions. 

The EMD, also known as the Wasserstein distance, measures the minimum cost of transporting the predicted point distribution to exactly match the ground-truth distribution.
\begin{equation}
\text{EMD}(\hat{\mathcal{P}}, \mathcal{P}) = 
\min_{\phi : \hat{\mathcal{P}} \rightarrow \mathcal{P}} 
\frac{1}{|\hat{\mathcal{P}}|} \sum_{p \in \hat{\mathcal{P}}} \| p - \phi(p) \|_2,
\label{eq:emd}
\end{equation}
Unlike Chamfer Distance, which matches each point independently to its nearest neighbor, EMD enforces a one-to-one global correspondence, thereby capturing both local geometry and global structure consistency.

\begin{table}[ht]
    \centering
    \setlength\tabcolsep{3pt}
    \renewcommand\arraystretch{1}
    \vspace{-1em}
    \caption{Comparison with state-of-the-art methods in F-score@$\tau$ ($\tau=0.001$) on ShapeNet In-Context~\cite{fang2023PIC}.}
    \vspace{-1em}
    \begin{tabular}{c|ccc}
        \toprule[0.15em]
        {\multirow{2}{*}{Models}}  & {Rec.}  & {Den.} & {Reg.}  \\ 
        {}                         & {F-score@$\tau$}   & {F-score@$\tau$}  & {F-score@$\tau$}                          \\\midrule
        PIC-Cat~\cite{fang2023PIC} & 38.36   & 33.06  & 26.98     \\
        PIC-Sep~\cite{fang2023PIC} & 42.78    & 32.86  & 41.57     \\
        \rowcolor{mypink}
        DeformPIC-\textit{8d}      & \textbf{57.98}    & \textbf{49.95}  & \textbf{66.04}
        \\\bottomrule[0.1em]
    \end{tabular}
    \vspace{-1em}
    \label{tab:more_metric_f_score}
\end{table}

In the F-score metrics (Table~\ref{tab:more_metric_f_score}), DeformPIC-\textit{8d} achieves the best performance on all three tasks (reconstruction, denoising, and registration), surpassing PIC-Cat and PIC-Sep by large margins. Notably, its registration F-score reaches 66.04, which is 39.06 points higher than PIC-Cat and 24.47 higher than PIC-Sep, indicating substantially improved robustness under large geometric perturbations. A similar trend appears in reconstruction and denoising, where DeformPIC-\textit{8d} outperforms baselines by 15–20 points, highlighting its ability to model both global structures and fine-grained local variations.

\begin{table}[ht]
    \centering
    \setlength\tabcolsep{3pt}
    \renewcommand\arraystretch{1}
    \vspace{-1em}
    \caption{Comparison with state-of-the-art methods in EMD ($\times1000$) on ShapeNet In-Context~\cite{fang2023PIC}.}
    \vspace{-1em}
    \begin{tabular}{c|ccc}
        \toprule[0.15em]
        Models                     & {Rec. EMD}  & {Den. EMD} & {Reg. EMD}  \\ \midrule
        PIC-Cat~\cite{fang2023PIC} & 56.30   & 52.89  & 75.14     \\
        PIC-Sep~\cite{fang2023PIC} & 60.83    & 65.27  & 64.28     \\
        \rowcolor{mypink}
        DeformPIC-\textit{8d}      & \textbf{37.01}    & \textbf{47.32}  & \textbf{24.05}  
        \\\bottomrule[0.1em]
    \end{tabular}
    \vspace{-1em}
    \label{tab:more_metric_emd}
\end{table}

In the EMD metrics (Table~\ref{tab:more_metric_emd}, DeformPIC-\textit{8d} again exhibits strong performance arcoss three tasks, achieving the lowest EMD values (37.01, 47.32 and 24.05, respectively), reflecting more accurate geometric preserving ability. Overall, the results indicate that DeformPIC-\textit{8d} generalizes significantly better in challenging in-context settings, particularly when structural consistency and alignment accuracy matter most.

\section{More Analysis}
\label{sec:supp_more_analysis}
Our DEN aims to extract task features that encode both task semantics (i.e., \emph{what} the task is) and transformation cues (i.e., \emph{how} to transform). To verify this, we construct a \emph{static} DEN variant that is supervised to predict a fixed task ID.
\begin{table}[h]
\centering
\vspace{-1em}
\caption{Static vs. Dynamic Encoding on DEN.}
\vspace{-1em}
\label{tab:static_den}
    \resizebox{0.48\textwidth}{!}{
    \large
    \begin{tabular}{c|ccc}
        \toprule[0.15em]
        Encoding Strategies    & Rec. CD $\downarrow$ &  Den. CD $\downarrow$ & Reg.CD $\downarrow$  \\
        \midrule
        Static ({\small Predict Task ID})      & 3.2  & 4.0 & 9.9  \\
        \rowcolor{mypink}
        Dynamic ({\small Predict Task Token})    & \textbf{2.7}       & \textbf{3.5}       & \textbf{2.0}  \\
        \bottomrule[0.1em]
    \end{tabular}
    }
    \vspace{-1em}
\end{table}

As shown in Table~\ref{tab:static_den}, static encoding results in a clear performance drop, suggesting that task-level information alone is insufficient. In contrast, the dynamic DEN captures prompt-conditioned, instance-specific deformation cues, which are crucial for effective geometric transfer.

\section{More Visualization}
\label{sec:supp_more_visualization}
\begin{figure*}[t]
    \centering
    \includegraphics[width=1.0\linewidth]{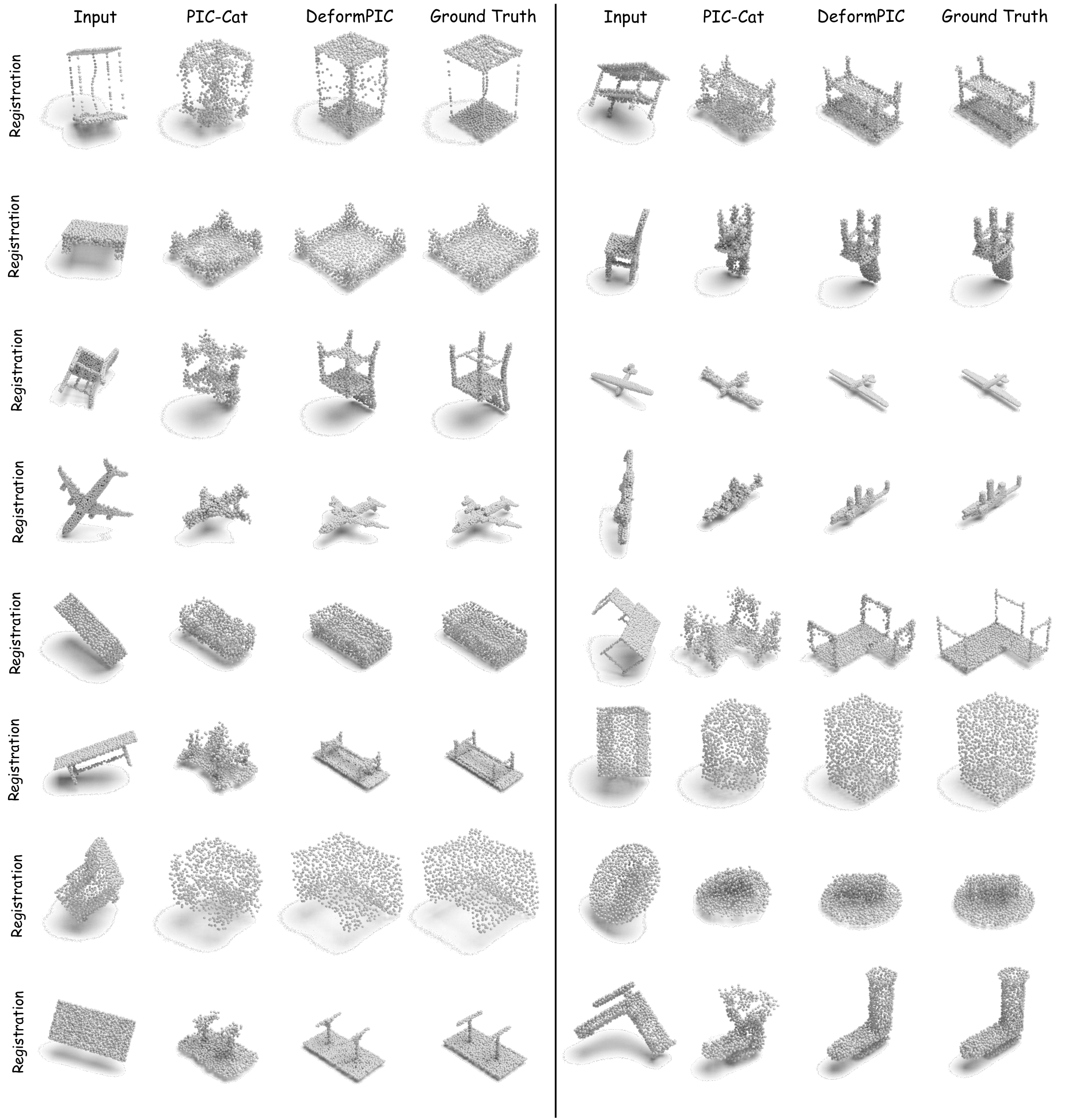}
    \caption{\textbf{Visualization results on registration task compared with the PIC~\cite{fang2023PIC} and our approach.}}
    \label{fig:supp_visualization_registration}
\end{figure*}

\begin{figure*}[t]
    \centering
    \includegraphics[width=1.0\linewidth]{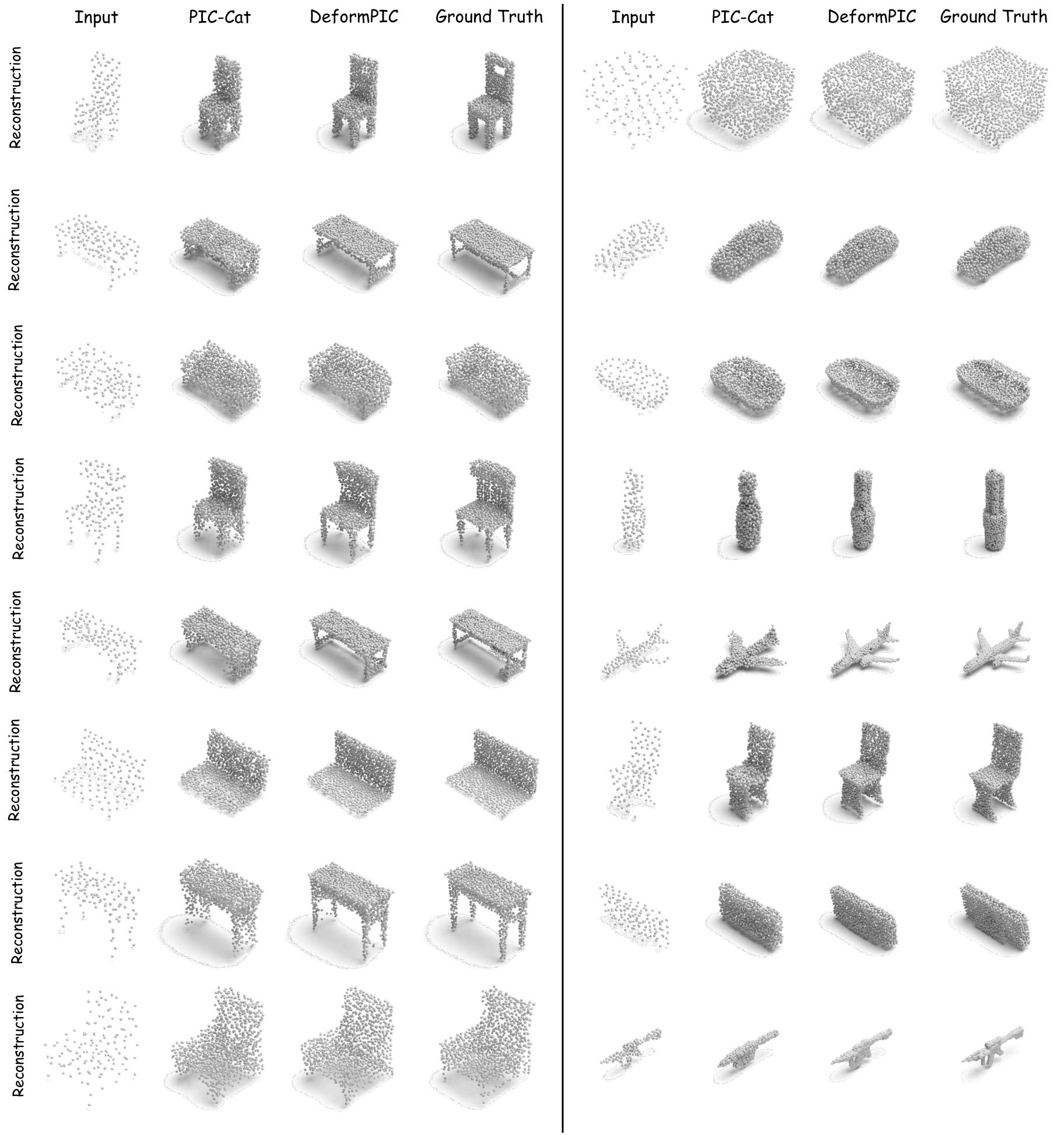}
    \caption{\textbf{Visualization results on reconstruction task compared with the PIC~\cite{fang2023PIC} and our approach.}}
    \label{fig:supp_visualization_reconstruction}
\end{figure*}

\begin{figure*}[t]
    \centering
    \includegraphics[width=1.0\linewidth]{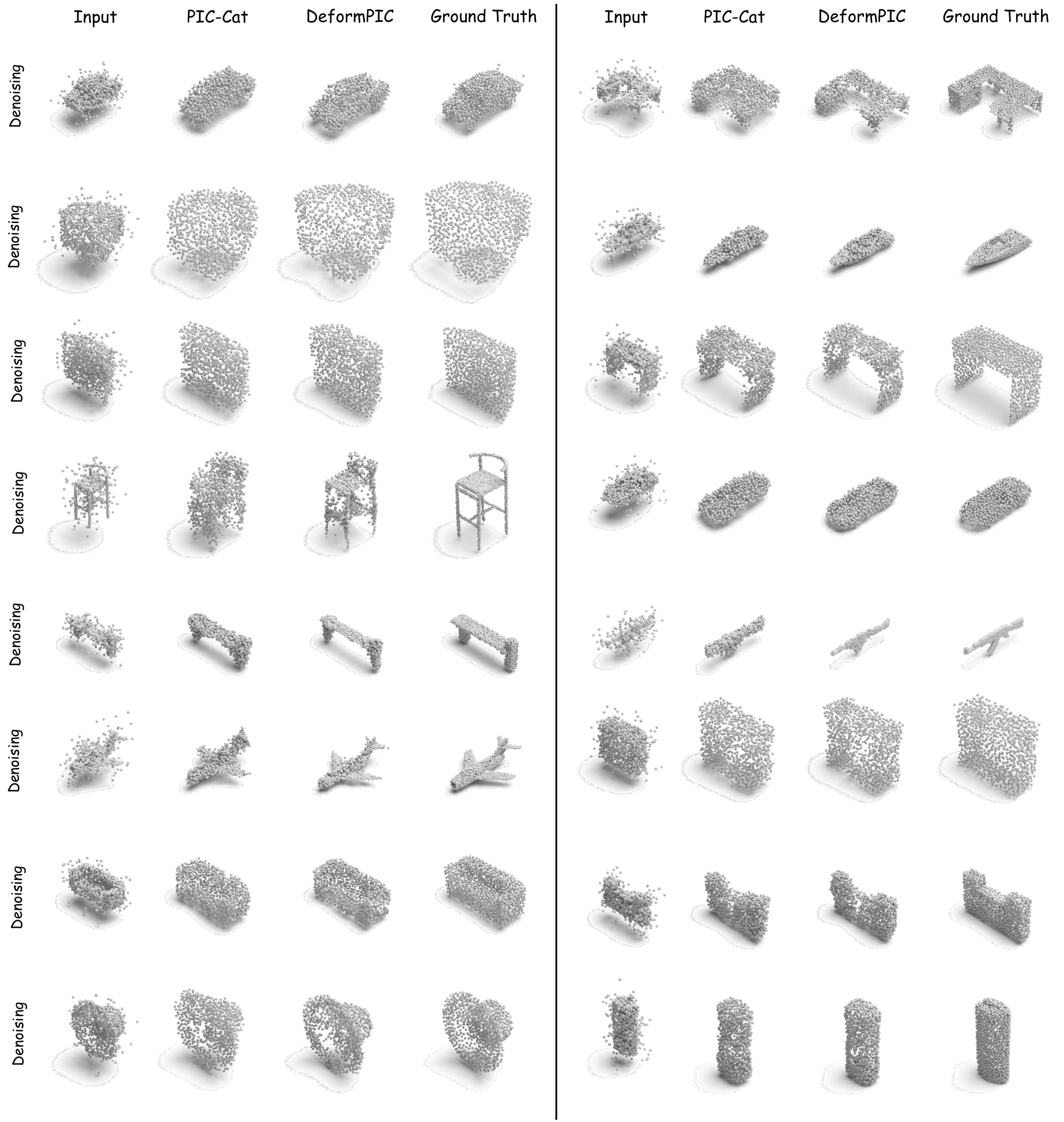}
    \caption{\textbf{Visualization results on denoising task compared with the PIC~\cite{fang2023PIC} and our approach.}}
    \label{fig:supp_visualization_denoising}
\end{figure*}

\begin{figure*}[t]
    \centering
    \includegraphics[width=1.0\linewidth]{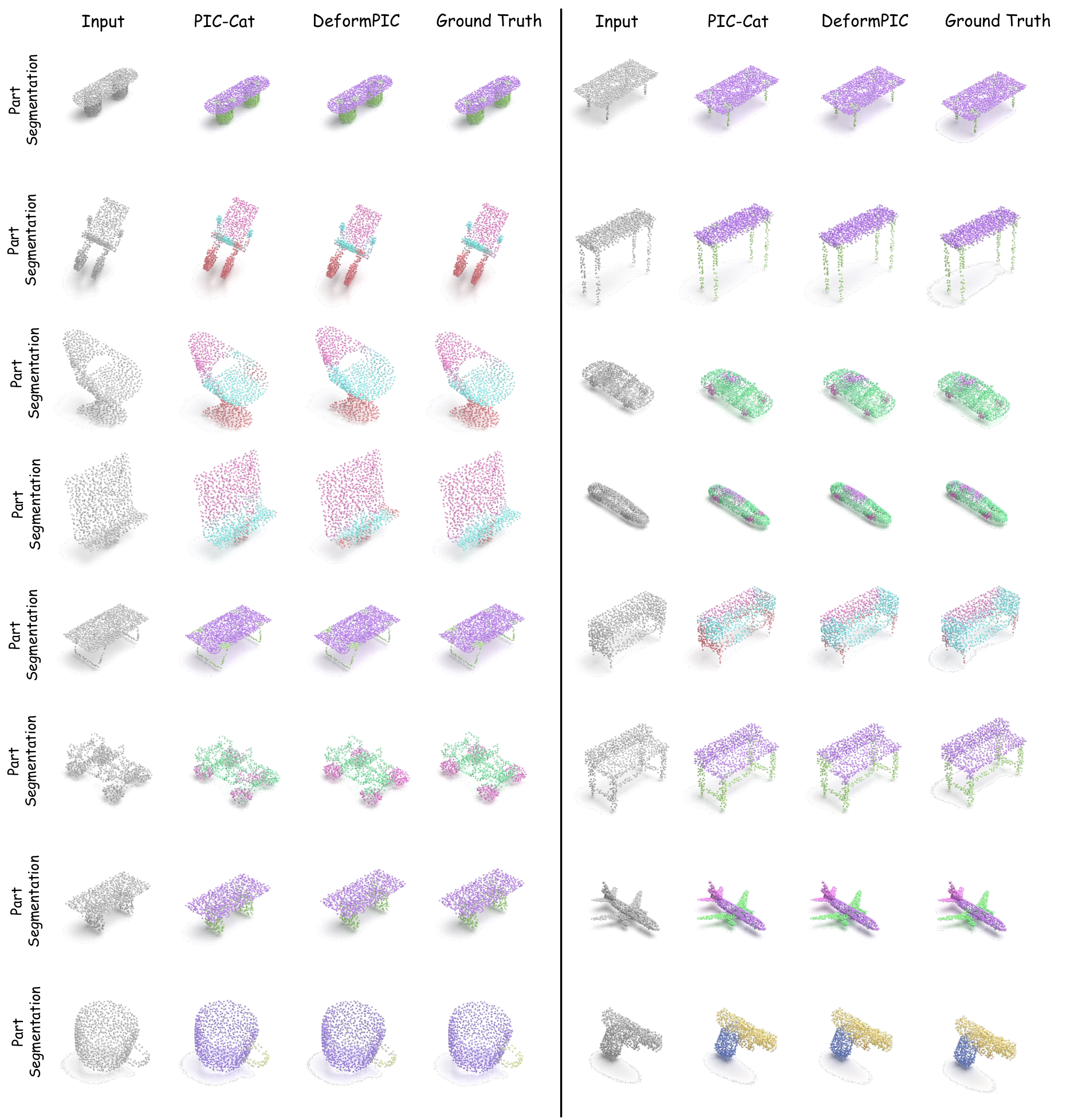}
    \caption{\textbf{Visualization results on part segmentation task compared with the PIC~\cite{fang2023PIC} and our approach.}}
    \label{fig:supp_visualization_partseg}
\end{figure*}

In this section, we provide more qualitative comparisons of our DeformPIC with PIC-Cat~\cite{fang2023PIC} on ShapeNet In-Context in Fig.~\ref{fig:supp_visualization_registration},~\ref{fig:supp_visualization_reconstruction},~\ref{fig:supp_visualization_denoising},~\ref{fig:supp_visualization_partseg}. Across all tasks, the results indicate that PIC-Cat~\cite{fang2023PIC} frequently exhibits geometric inconsistencies with respect to the query point cloud. In some cases, it even produces noticeable shape distortions. These observations further highlight the strong geometric preservation ability of our DeformPIC model.

\section{Discussion}
\label{sec:supp_discussion}
\noindent\textbf{Limitations.}
Although our method achieves excellent performance on out-of-domain data, it is still premature to consider it a general-purpose model for point cloud understanding, given that its capabilities are limited to four tasks: reconstruction, denoising, registration, and part segmentation. In addition, current point cloud ICL techniques operate only on object level point clouds, and it remains unclear whether they can be effectively extended to indoor or outdoor scene level data.

\noindent\textbf{Future Works.}
We hypothesize that a key factor underlying the success of ICL in LLMs is the availability of large-scale training data and diverse task types. Motivated by this, our future work will investigate training point cloud ICL models on substantially larger and more varied datasets to move toward truly point cloud understanding generalist. In addition, we will explore applying these models to specific scene level settings, such as embodied perception and autonomous driving.

\end{document}